\documentclass[runningheads]{llncs}
\usepackage{graphicx}
\usepackage{comment}
\usepackage{caption}
\usepackage{subcaption}
\begin{document}
\title{A Deep Hybrid Model for Recommendation Systems}
%
%
\author{Muhammet \c{C}ak{\i}r\inst{1} \and
\c{S}ule G\"{u}nd\"{u}z \"{O}\u{g}\"{u}d\"{u}c\"{u}\inst{1} \and
Resul Tugay\inst{1}}
\authorrunning{M. \c{C}ak{\i}r et al.}
%
\institute{Computer Engineering Department, Istanbul Technical University, Istanbul, Turkey \\
\email{\{cakirmuha,sgunduz,tugayr\}@itu.edu.tr}}
\maketitle              
\begin{abstract}
Recommendation has been a long-standing problem in many areas ranging from e-commerce to social websites. Most current studies focus only on traditional approaches such as content-based or collaborative filtering while there are relatively fewer studies in hybrid recommender systems. Due to the latest advances of deep learning achieved in different fields including computer vision and natural language processing, deep learning has also gained much attention in Recommendation Systems. There are several studies that utilize ID embeddings of users and items to implement collaborative filtering with deep neural networks. However, such studies do not take advantage of other categorical or continuous features of inputs. In this paper, we propose a new deep neural network architecture which consists of not only ID embeddings but also auxiliary information such as features of job postings and candidates for job recommendation system which is a reciprocal recommendation system. Experimental results on the dataset from a job-site show that the proposed method improves recommendation results over deep learning models utilizing ID embeddings.

\keywords{Content-based Filtering  \and Collaborative Filtering \and Hybrid Systems \and Deep Neural Networks \and Job Recommendation\and Implicit Feedback.}
\end{abstract}
\section{Introduction}

Recommendation is the problem of predicting the ratio of interaction between users and items such that a user would prefer an item to another one. It is a ubiquitous problem that appears in numerous application domains ranging from dating websites to e-commerce websites. Recommendation systems (RSs) make web experience special to their users by recommending, what to buy (e-commerce, Amazon, Ebay), which movies to watch (Netflix), who to follow (Twitter), which songs to listen (Spotify) etc. We mainly focus on a special type of RSs in this paper, Job Recommendation (JR), which is different than conventional RSs with various aspects. In reciprocal RSs, preferences of both users and items should be considered to find a suitable match and generate recommendations as different from traditional RSs, where items are recommended to users. In a reciprocal RS for a job recruitment web service, jobs are recommended to applicants on one side and  candidates are also recommended to recruiters on the other side. The recommendation is successful only if the preferences of users at both sides are satisfied. Moreover, there are mostly no explicit feedback for recruiters and job applicants in JR as there are ratings, likes/unlikes etc. for items in conventional RSs. 

In general, recommendation lists are created based on user-item past interactions, item properties, user preferences, and some other additional data. There are three main techniques used in traditional RSs: Content Based Filtering (CBF), Collaborative Filtering (CF) and Hybrid RSs.

\textbf{Content-Based Filtering:} This technique creates user profiles taking which users previously interact with items into account and simply recommends items with similar contents to user profiles. The recommendation process uses properties of items as contents of users \cite{mooney2000content,pazzani2007content,lops2011content}.

\textbf{Collaborative Filtering:} 
The CF based recommendation systems represent users' preferences as $n$-dimensional rating vectors where $n$ is the number of items in the system. The key idea of CF is that similar users/items share similar interests. \cite{sarwar2001item,schafer2007collaborative}. CF recommends items to users based on their liked items by computing similarities between users and items. There are two categories of CF: 
\begin{itemize}
    \item \textbf{Item-based:} Calculates the similarity between the items that user rates previously and other items.
    \item \textbf{User-based:} Calculates the similarity between users.
\end{itemize}

\textbf{Hybrid Systems:} These systems combine two or more types of recommendation techniques to produce better recommendations \cite{burke2002hybrid,burke2007hybrid}.

More recently, Deep Learning (DL) has gained a tremendous success which takes advantage of deep neural networks using contextual, textual, and/or visual information to produce better recommendation results. DL techniques can be used with all approaches to leverage existing recommendation system. On the other hand, these approaches can also be used not only in conventional RSs but also in JR which requires a special attention to adopt. In the past few years, DL has become a major direction in many fields including machine learning  \cite{goodfellow2016deep,schmidhuber2015deep} and RSs\cite{he2017neural,wang2015collaborative} etc. DL has been used to model users and items considering their properties as input of deep neural network. For instance, Oord \textit{et al.} used  time-frequency representation from the audio signals as input to the network \cite{van2013deep}. Zhang \textit{et al.} applied embedding using  items’ textual content and visual content to create semantic representation on deep network \cite{zhang2016collaborative}. Then, He \textit{et al.} also used another embedding technique which is called ID embedding that takes users and items IDs to make a good recommendation instead of using simple approach that applies an inner product on the latent features of users and items \cite{he2017neural}. We benefit from this ID embedding approach along with explicit features of job applicants and job postings to produce better recommendations in this paper.

Matrix Factorization (MF) \cite{koren2009matrix} which discovers hidden factors being the underlying reason for the user feedback is the most popular CF technique among the various approaches, and it will be expanded with ID embeddings in this work. Also, Deep Neural Networks(DNNs) \cite{covington2016deep} that is a popular CBF approach will be adopted to model auxiliary information. This paper proposes a hybrid system that utilizes both MF and DNNs; DNN joins content information into a collaborative approach with ID embeddings as the first part of the hybrid model, and the main framework combines collaborative latent features from MF, and the mixed features from DNN over the last interaction layer as the second part of the hybrid model.


The main contributions of this work are as follows:

\begin{enumerate}
    \item We propose a hybrid model that combines features of candidates and job postings along with ID embeddings. It joins the strengths of linearity of MF and non-linearity of DNN to model user $\times$ item interaction.
    \item A novel approach is built for reciprocal recommendation which can be easily applied on job recruitment websites. All experiments are performed with a real-world dataset provided by one of the biggest job-recruitment website in {{Turkey}}.
\end{enumerate}

The rest of the paper is organized as follows: Section \ref{sec:RW}
reports related work about Recommendation
Systems, Section \ref{sec:PR} provides preliminaries about RSs  Section \ref{sec:Model} presents our proposed method and general neural network architecture. In Section \ref{sec:Exp}, we show
the experimental evaluation presenting the methodology and
results. Finally, Section \ref{sec:Conclusion} provides conclusion and discusses some of possible future extensions that are special for JR. 

\section{Related Work}
\label{sec:RW}

Although there are some studies that focus on CBF \cite{lops2011content,van2013deep}, the majority of the approaches are based on CF \cite{koren2009matrix,lee2001algorithms,mnih2008probabilistic}. CF approaches mainly utilize feedbacks of users for items. Users’ feedbacks can be categorized in two ways: implicit feedback \cite{hu2008collaborative,he2016fast} which indirectly reflects preferences of users, and explicit feedback \cite{koren2008factorization} which directly indicates users’ choices. While implicit feedback shows behaviours like purchasing products, clicking items, and watching videos, users’ ratings and reviews for products are considered as explicit feedback \cite{he2017neural}. While it is difficult to make use of implicit feedback because of not observing user satisfaction, explicit feedback provides negative inference with low ratings and unfavorable reviews.

The original MF approach \cite{koren2009matrix} was proposed to model explicit feedbacks of users by mapping users and items to a latent factor space so that interactions (e.g. ratings) of users and items can be represented by dot product of latent factors of them. Many approaches have been presented to expand MF such as combining it with neighbor-based models \cite{koren2008factorization}, and extending it to factorization machines \cite{rendle2010factorization} for generating a model using user and item features. However, users mostly interact with items through implicit feedback since explicit ratings are not present all the time for many recommendations \cite{he2016fast}. In JRs, there is no explicit feedback like rating and like/dislike, but it is easy to collect implicit feedback such as job view and job application for content providers even being more challenging to use because of the natural lack of negative feedback. 

CBF is mainly based on comparisons across supporting information of users and items while CF approaches use only user-item interactions. Texts, images, and videos can be thought as a wide variety of additional information \cite{zhang2019deep}. At the last few years, many research efforts have been made to improve the recommendation performance with use of auxiliary information, and deep learning techniques have gained importance to process a great amount of information. They generally adopted on DNNs for modeling auxiliary information using implicit feedback, such as content of videos \cite{covington2016deep}, audible features of musics \cite{van2013deep,wang2014improving}, textual explanation of items \cite{wang2015collaborative}, and categorical-continuous features of both users and items \cite{cheng2016wide}. Although many applications like music and video websites have no more personal information for users and limited amount of information such as textual description, job search websites fortunately contain many categorical and continuous features for both user and item sides to adopt DL approaches.

There have been a great number of hybrid approaches proposed by combining collaborative and content-based methods \cite{albadvi2009hybrid,balabanovic1997fab,soboroff1999combining}. A hybrid approach can be implemented in different ways such as applying separately collaborative and content-based tasks and combining their predictions, joining some content-based properties into a collaborative approach, or opposite, and lastly building a general mixed model that joins collaborative and content-based
properties \cite{adomavicius2005toward}. In this work, we will adopt on both a general fixed model and concatenating content-based properties with a collaborative approach.

Furthermore, there are no more general approaches for implementing JR task. Some studies \cite{wang2017dionysius,lee2016job} particularly focus on JR using a specific dataset such as modeling the interactions for user-job record, user-company, user-job title, and recommendation through graph analysis in AskStory \cite{lee2016job}, and incorporating user interactions into the recommendation task with a hierarchical graphical model in LinkedIn \cite{wang2017dionysius}. In contrast to these models, our proposed model aims at bringing out a framework that is available for many applications such as book, music, and movie recommendation while making better recommendation for both job seekers and recruiters in JR. Also, it may not be reasonable to make reciprocal recommendation for many real-world applications since there is no need to recommend a user for an item like user recommendation for books, movies, and musics.
However, some studies \cite{li2012meet,xia2015reciprocal} try to build a generalized framework for generating the list of the most suitable candidates. The elements of reciprocal recommendation can be two users having bilateral relations as traditional user $\times$ item interactions. Xia \textit{et al.} brings a solution to online dating problem by finding out users' power of relationships analysing user-based and graph-based features \cite{xia2015reciprocal}. Reciprocal recommenders, such as online dating and online recruiting, have a major challenge to satisfy the mutual preferences of two users or user $\times$ item, and graph-partitioning methods are used by representing the dataset as a bipartite graph \cite{li2012meet}. While these works make a good job of graph analysis to find out the relationships of users, deep learning based approaches has been disregarded to process a great amount of information existing in properties of users and items, textual explanation, and video/music content. 

\section{Preliminaries}
\label{sec:PR}

In this part of the paper, existing solutions for CF and CBF  that inspire Hybrid Recommender System (HRS) will be mentioned. These preliminaries including Matrix Factorization (MF) that utilizes user-item IDs and Deep Neural Network (DNN) that takes advantages of user-item properties build the main structure of our model.

\subsection{Matrix Factorization}

MF technique is only used for CF, and allows to discover the latent features underlying the interactions between users and items. Each user and item can be associated with a real-valued vector of the latent features. Let $p_u$ and $q_i$ denote the latent vectors for user $u$ and item $i$, respectively; MF is used to estimate an interaction $\hat{y}_{ui}$ as the dot product of $p_u$ and $q_i$:

\begin{equation}
    \hat{y}_{ui} = f(u, i|p_{u},q_{i}) = \sum_{k=1}^{K} p_{uk}q_{ki}  
\end{equation}
where $K$ is a parameter that represents the dimension of the latent space. MF models the interaction of user and item latent factors, and each dimension of the latent space is independent from each other. Dimensions of the latent space are linearly combined with the same weight, so MF can be considered as a linear model of latent factors.

\subsection{Deep Neural Networks}

There are a variety of deep learning based approaches for CF and CBF. In this part, Multi-layer perceptrons (MLP) that is a type of feedforward artificial neural network will be mentioned. User and item properties are integrated by concatenating them. Interaction between user and item latent features does not exclusively occur by a vector concatenation that is unsatisfactory for CF modeling \cite{he2017neural}. This problem is resolved by adding some hidden layers after concatenating these latent vectors.  Let $p_u$ and $q_i$ denote the latent vectors for user $u$ and item $i$, respectively; DNN model is simply formulated as

$$
\psi_{1}(p_{u}, q_{i}) = o_1 =
\bigg [
  \begin{tabular}{c}
  $p_u$ \\ 
   $q_i$
  \end{tabular}
\bigg ]
$$

\begin{figure*}[ht!]
\centering
  \includegraphics[width=\linewidth] {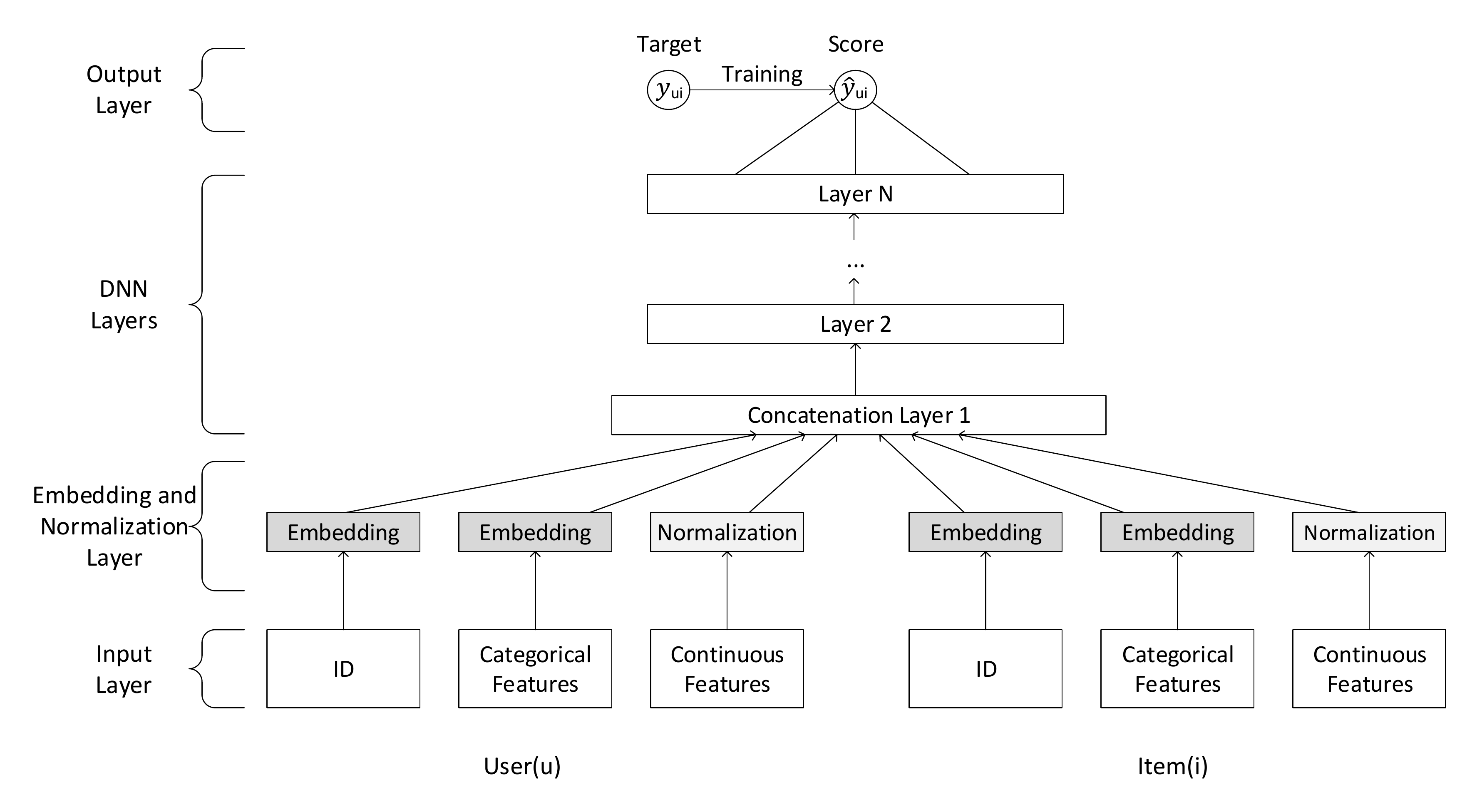}
  \caption{General Architecture of Deep Neural Network}
  \label{fig:DNN}
\end{figure*}

$$ \psi_{2}(o_{1}) = o_2 = \alpha_{2}(W_2^To_{1} + b_2),$$ 
$$ ...$$ 
$$  \psi_{N}(o_{N-1}) = o_N = \alpha_{L}(W_N^To_{N-1} + b_N),$$ 
\begin{equation}
    \label{eqn:DNN}
     \hat{y}_{ui} = \alpha_{out}(h^To_{N})
\end{equation}
where $W_l$ and $b_l$ specify the weight matrix, bias vector, respectively. $\alpha_l$ is the activation function of the $l$-th layer of the model. Also, $\psi_l$ represents a function that takes the output of $l-1$-th layer as input, and produces the result of the current layer.

\section{DeepHybrid Model}
\label{sec:Model}

In this paper, a hybrid recommender system is proposed using user and item IDs for user-item historical interactions, and auxiliary information of users and items. This model is called as DeepHybrid. Firstly, latent features of users and items are obtained using MF technique from feedback of users. Secondly, a deep learning based approach is used to get users' and items' features from both user-item IDs and user-item properties. Lastly, a hybrid model is presented, and it combines MF and DNN under DeepHybrid.

\subsection{Features from MF}

MF is the most popular approach for CF, and it maps users and items to a latent factor space for the interaction of users and items. 
In this work, $p_u$ represents user properties and $q_i$ represents item properties. Embedding vector with one-hot encoding of user (item) ID of the input layer can be seen as the latent vector of user (item) \cite{he2017neural}. Let the user latent vector $p_u$ be $P_Tv_u^U$ and item latent vector $q_i$ be $Q_Tv_i^I$. These latent vectors are projected to the output layer as
\begin{equation}
    \hat{y}_{ui} = \alpha_{out}(h^T(p_{u} \odot q_{i})) 
\end{equation}
where $\odot$ defines the element-wise product of vectors, and $\alpha_{out}$ represents the activation function. Also, $h$ is used as weight vector of the output layer.
In our framework, MF model will be expanded using an activation function. Unlike linear MF models, a non-linear activation function will make our MF model more meaningful.

\subsection{Features from DNN}

In our DNN model, auxiliary information of users and items will be added for concatenated vector that is the input of the neural network. In addition to embedding vectors obtained from one-hot encoding of user and item IDs like MF, the properties of users and items will be processed as embedding vector or normalized value. These properties can be divided into two categories that are continuous features and categorical features. Firstly, continuous features are real-valued numbers and normalized to [0, 1] by mapping a feature value $x$. The normalized value is calculated as Equation \ref{eqn:normalization}  for values in the $i$-th quantiles. 
\begin{equation}
    \label{eqn:normalization}
    n_i = \frac{i-min}{max-min}
\end{equation}

Secondly, categorical features are embedded to $n$-dimensional vectors like ID embeddings depending on the number of distinct values for the stated feature. For example, age of a candidate is a continuous feature and normalized into [0, 1], and military status is considered as a categorical feature since this value is distinct, and simply divided into two as completed and uncompleted statuses in JRs. Figure \ref{fig:DNN} shows the architecture of DNN to process features after embedding and normalization. 
Let $con_u$, $cat_u$, and $id_u$ represent continuous features, categorical features and ID embedding of users, respectively; and similar notations of $con_i$, $cat_i$, and $id_i$ for items. User and item properties for DNN model is simply defined as

\begin{equation}
p_u = \Bigg[\begin{tabular}{c}$id_u$ \\$cat_u$ \\$con_u$ \end{tabular} \Bigg], q_i = \Bigg[\begin{tabular}{c}$id_i$ \\$cat_i$ \\$con_i$ \end{tabular} \Bigg] 
\end{equation}

These properties of users and items are concatenated after embeddings and normalization. Then, the result features of these operations are processed by DNN as Equation \ref{eqn:DNN}. In this manner, the model learns the interaction between $p_u$ and $q_i$ by providing more flexibility and non-linearity than MF that operates only a dot product of user and item latent vectors.

\subsection{Interaction Layer}
\begin{figure}[ht!]
\centering
  \includegraphics[width=\linewidth] {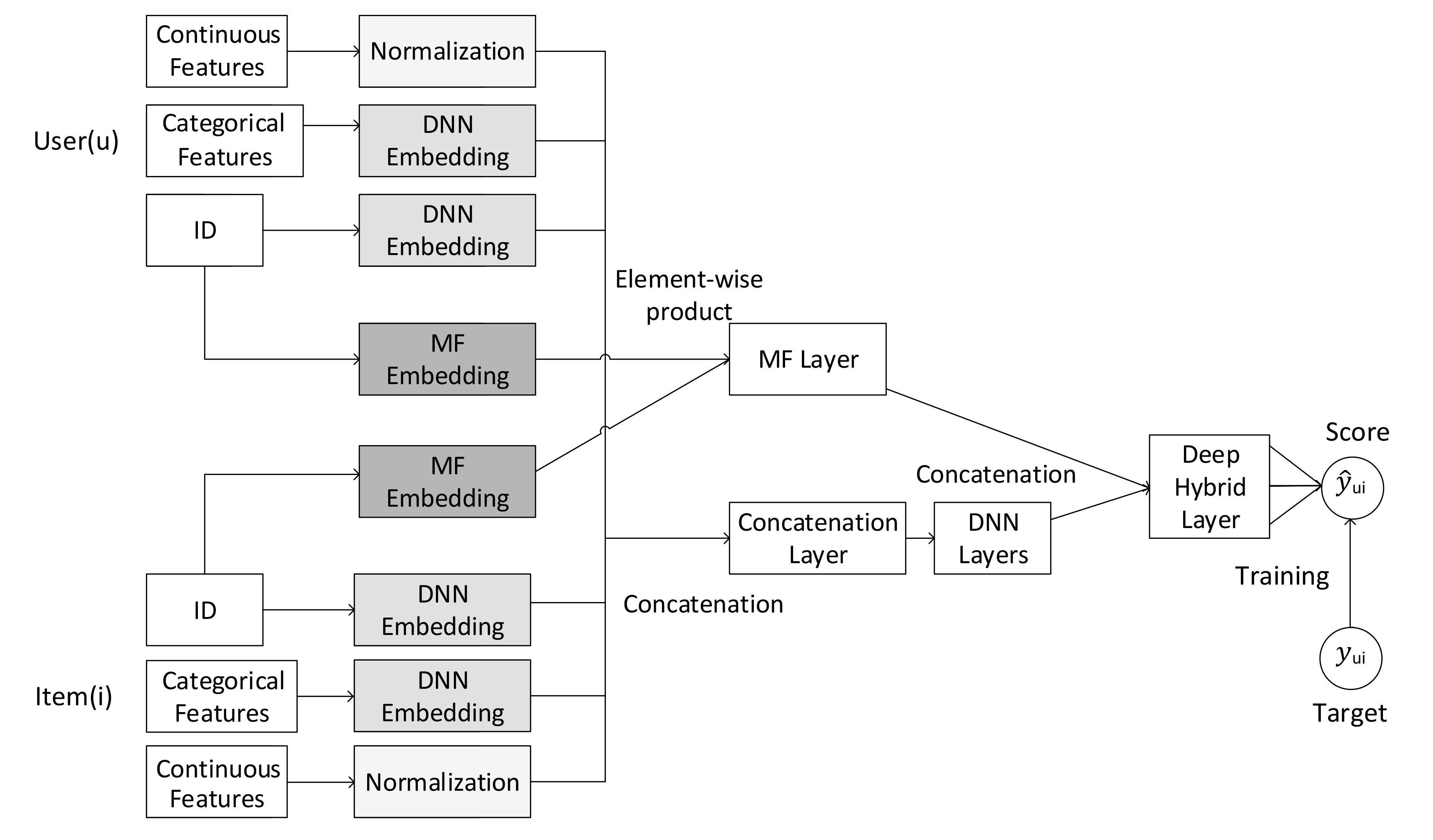}
  \caption{DeepHybrid Model}
  \label{fig:DHM}
\end{figure}

In the interaction layer, two different mappings of latent features from categorical, continuous features, and ID embeddings are fused in DeepHybrid as shown in figure \ref{fig:DHM}. This model basically aims at ranking prediction as recommendation task. A top-n list is produced by the ranking network using the nearest neighbors of the candidates \cite{zhang2019deep}.
MF and DNN models can share the same embedding layers for processing user and item IDs, then concatenate the results of their operations. However, He \textit{et al.} stated that MF and DNN use different embeddings, and are fused by concatenating the last layers of these models to enable more flexibility to the combined model \cite{he2017neural}. 

$$ \psi^{MF} = p_u \odot q_i$$
$$
\psi^{DNN} = \alpha_N(W_N^T(...\alpha_2(W_2^T\bigg[\begin{tabular}{c}$p_u$ \\$q_i$\end{tabular} \bigg] + b_2)...)) + b_N 
$$

\begin{equation}
    \hat{y}_{ui} = \alpha_{out}(h^T\bigg[\begin{tabular}{c}$\psi^{MF}$ \\$\psi^{DNN}$\end{tabular} \bigg])
\end{equation}

\subsection{Learning Interaction}
An objective function needs to be specified to estimate model parameters. Existing solutions generally utilize a regression with squared loss. For example, rating prediction is originally a regression problem, and squared loss will exactly fit rating prediction. However, our target value $y_{ui}$ has only two values 1 or 0 representing whether $u$ and $i$ have an interaction. 

DeepHybrid uses the binary property of data, and $y_{ui}$ is labeled as 1 that means user $u$ has positive feedback on item $i$, and 0 otherwise. The prediction score $\hat{y}_{ui}$ shows how user $u$ is related to item $i$. In this point, our recommendation task with implicit feedback can be considered as a binary classification problem. If the probability of class-1 is high, $u$ is more relevant to $i$, or vice versa. Let $y_{ui}$ and $\hat{y}_{ui}$ denote the target and the predicted value, \textbf{X} be the interaction of users and items, respectively; 

\begin{equation}
\label{eqn:logloss}
     L = \sum_{(u,i)\in X\cup X^-} y_{ui} \log \hat{y}_{ui} + (1 - y_{ui}) \log (1 - \hat{y}_{ui})
\end{equation}
where $X$ denotes the set of observed interactions in \textbf{X}, and $X^-$ denotes
unobserved interaction. This formula is called as \textit{log loss} (\textit{binary cross-entropy loss}). While $X$ represents positive instances, $X^-$ is determined by uniformly sampling from unobserved interactions due to a lack of negative feedback. For each positive instance, a certain number of negative instances are determined.

\section{Experiments}
\label{sec:Exp}

\subsection{Dataset and Experimental Settings}
In our experiments, we focused on Job Recommendation, and used the dataset of one of the biggest job search website in {{Turkey}}.

\textbf{Kariyernet:} Kariyernet dataset includes job application information like which candidate applied which job posting. Table \ref{table:stats} shows the statistics of this dataset. Also, auxiliary information of users and items are provided as:
\begin{itemize}
    \item \textbf{Candidate properties:} Firstly, age of a candidate is used as continuous feature, and normalized. Secondly, gender, military status, working status, city, department, and university are considered as categorical feature, and embedded.
    \item \textbf{Job posting properties:} The number of people to hire is used as continuous feature. All other features of job posting that include preferred gender, position type, position level, and educational status are taken into account as categorical feature.
\end{itemize}

Our proposed deep hybrid model was implemented on Keras using Theano backend. For all implemented models, one interaction of each user was sampled as the validation, then hyper-parameters were tuned. Negative instances were randomly selected from unobserved interactions because of lack of negative feedback. Three negative instances were allocated for each positive instance in training set. Also, model parameters were initialized with a Gaussian distribution, and optimized the binary cross entropy of equation \ref{eqn:logloss} with mini-batch Adam \cite{kingma2014adam}. For MF, the number of latent factors was chosen as 4,8,16, and 32 to obtain latent features from ID embeddings. For DNN, output dimension of embedding layer for ID embedding was selected as 16, 32, and 64. Also, normalization of continuous features naturally produces a number between 0 and 1, 
and categorical features were embedded to one-dimensional space. Moreover, the batch size for each iteration was tested with values 100, 200, and 400, and the learning rate was applied with 0.0001, and 0.001 for Adam optimizer. Lastly, our models showed better performance with latent factor as 16, input layer size for DNN as 64, batch size as 200, and learning rate as 0.001.

\begin{table}
\caption{Statistics of Kariyernet dataset}
\label{table:stats}
\centering
 \begin{tabular}{||c|c|c|c||} 
 \hline
 Candidates & Job postings & Applications & Sparsity \\ [0.5ex] 
 \hline\hline
 20283 & 16134 & 383434 & 99.99\% \\ [1ex] 
 \hline
\end{tabular}
\end{table}

\subsection{Evaluation Metric}
\textit{Leave-one-out} evaluation strategy was selected to evaluate both user recommendation and item recommendation. This evaluation that has been mostly used in other works \cite{he2016fast,rendle2009bpr} takes all data for training except the latest interaction of user or item as the test set. The ranking strategy generates a top-n list of items for each user. Also, m items are randomly sampled for each user as negative instances that is different from training phase. He \textit{et al.} uses \textit{Hit Ratio} (HR) and \textit{Normalized Discounted Cumulative Gain} (NDCG) to evaluate the performance of top-n list \cite{he2017neural}. This work randomly selects 99 items from unobserved items for each user in the test phase, and ranks top 10 of these items while implementing item recommendation, or the opposite for user recommendation. While NDCG is calculated by giving top items more higher score than next items, HR only accounts if the test item is available on top-10 item list. 

\subsection{Baseline Models and Performance Evaluation}

\subsubsection{Baselines}
Baseline models were selected from the studies which utilize interactions between users and items, and calculates top-n list of items.
\begin{itemize}
    \item \textbf{eALS(CF):} Matrix Factorization is the most popular recommendation model for CF. eALS forms a special type of MF with implicit feedback that handles all unobserved instances as negative, but gives weight to these instances depending on the popularity of items \cite{he2016fast}.
    \item \textbf{MF with ID Embedding(CF):} This method generalizes, and expands classical MF works. Unlike linear MF models, it takes ID embeddings of users and items, and predicts the target value using element-wise product with a nonlinear activation function like sigmoid, tanh etc \cite{he2017neural}.
    \item \textbf{DNN(CBF)}: This model takes only content information of users and items instead of IDs. The features are collected with the traditional taxonomy of continuous and categorical features. It embeds categorical features, normalizes continuous features, and  a neural network processes these concatenated features \cite{covington2016deep}.
    \item \textbf{NeuMF(CF):} This model is only designed for CF. It uses ID embeddings that are processed under GMF, and MLP, and then combined under an interaction layer by concatenating each other. The result is optimized by \textit{binary cross entropy loss} \cite{he2017neural}. 
\end{itemize}

\subsubsection{Performance Evaluation}
In JR systems, both jobs can be recommended to candidates, and the most perfectly fit resumes can be determined for each job posting. In this work, the results were calculated for both item and user recommendation. The performance of DeepHybrid and baseline models on Kariyernet dataset is shown with HR@10 and NDCG@10 as evaluation metric in Table \ref{table:dhm}. The experiments were repeated three times, and the averages of the results were written down in terms of evalution metrices HR and NDCG.

\begin{figure*}[ht!]    
    \centering
    \begin{subfigure}{0.46\textwidth}
        \includegraphics[width=\linewidth] {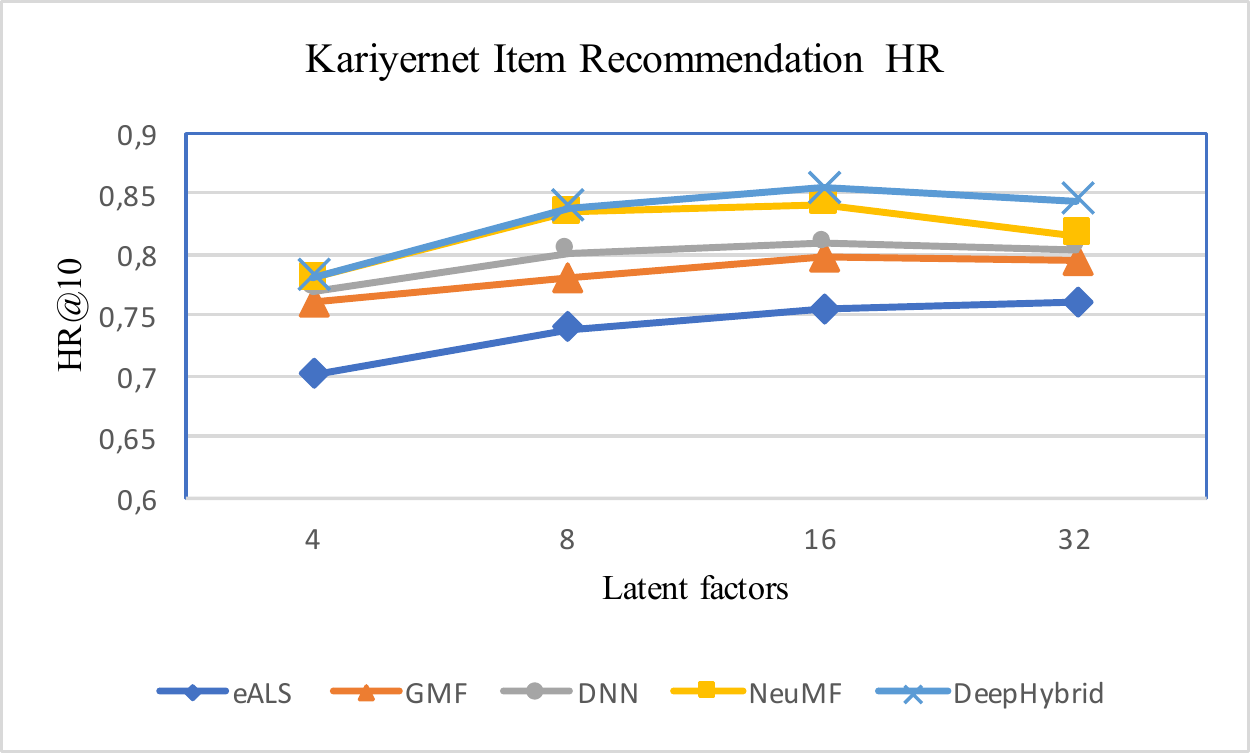}
        \caption{Factors vs HR@10}
    \end{subfigure}
    \hfill
    \begin{subfigure}{0.46\textwidth}
        \includegraphics[width=\linewidth] {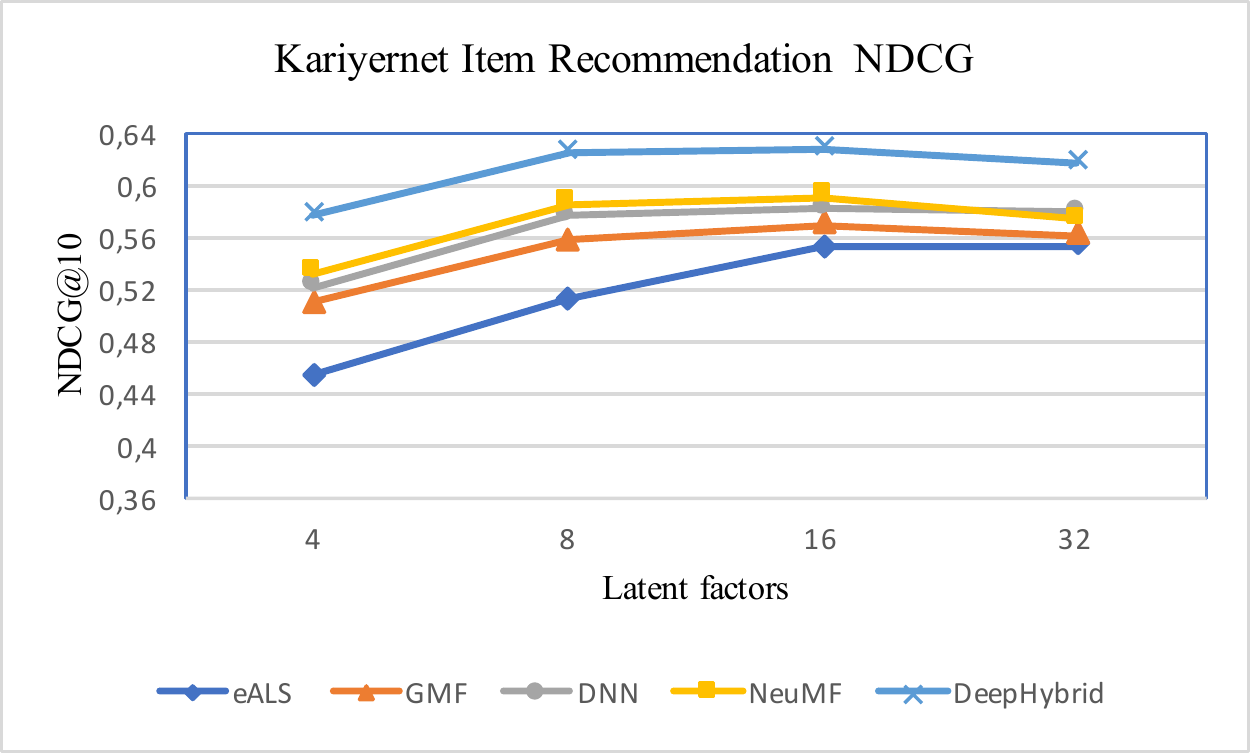}
        \caption{Factors vs NDCG@10}
    \end{subfigure}
    \caption{Performance of HR@10 and NDCG@10 with the number of factors}
    \label{fig:factors}
\end{figure*}

In this work, DeepHybrid achieves high improvements over the state-of-the-art methods on Kariyernet dataset. Firstly, GMF and eALS use only the interaction of user and item, and GMF importantly outperforms eALS that shows that the non-linearity setting of GMF is more expressive than linear MF models like eALS. Secondly, DNN contributes on increasing HR and NDCG using only content information that is stronger learner for ranking performance. Thirdly, NeuMF made a valuable improvement on the other baseline models for both item and user recommendation since it combines the linearity and non-linearity of models, and it provides more flexibility to learn the interaction between user and item latent factors using a deep learning based approach. These three baseline models except DNN that utilizes auxiliary information use only user-item historical interaction, and are called as CF approaches. DeepHybrid achieves the best performance over all baseline models by integrating CBF and CF strategies. This model concatenates categorical and continuous features into ID embeddings, and makes latent features of users and items stronger. DeepHybrid gains approximately \textit{\textbf{\%1.7}} improvement over NeuMF for both item and user recommendation in terms of HR. However, the results shows that the improvement of DeepHybrid is \textit{\textbf{\%6}} in terms of NDCG for Kariyernet dataset, and it can be said that properties of candidates and job postings provide a strong performance for ranking on top-n list.

\begin{table}
\caption{Performance of DeepHybrid Model}
\label{table:dhm}
\centering
 \begin{tabular}{|c|c|c|c|c|} 
 \hline
  & \multicolumn{2}{|c|}{\textbf{Item recommendation}} & \multicolumn{2}{|c|}{\textbf{User recommendation}}  \\
 \hline
 \textbf{Methods} & \textbf{HR@10} & \textbf{NDCG@10} & \textbf{HR@10} & \textbf{NDCG@10} \\ [0.5ex] 
 \hline\hline
 \textbf{eALS} & 0.754 & 0.566 & 0.672 & 0.431 \\ [1ex] 
 \hline
 \textbf{GMF} & 0.797 & 0.569 & 0.752 & 0.491 \\ [1ex] 
 \hline
 \textbf{DNN} & 0.809 & 0.582 & 0.761 & 0.524 \\ [1ex] 
 \hline
 \textbf{NeuMF} & 0.841 & 0.591 & 0.779 & 0.543 \\ [1ex] 
 \hline
 \textbf{DeepHybrid} & \textbf{0.855} & \textbf{0.628} & \textbf{0.795} & \textbf{0.581} \\ [1ex] 
 \hline
\end{tabular}
\end{table}

Also, figure \ref{fig:factors} shows the performance of HR@10 and NDCG@10 depending on the number of latent factors. While this factor represents the latent vector size for eALS and GMF, it is considered as the output of the last layer for other DNN based approaches. For all models, HR and NDCG evaluation metrices have lower values for latent factor 4, but they are so close to each other for 8, 16, 32. These results showed us that latent factor 16 is the most suitable one to evaluate all models since large factors cause overfitting while small factors are not sufficiently learning the model depending on the size of dataset.

\section{Conclusion}
\label{sec:Conclusion}
In this work, a deep learning based hybrid recommender system is proposed. It utilizes user-item interaction for CF and auxiliary information of both users and items for CBF. DeepHybrid consists of MF and DNN coupled together by a shared common layer to predict top-n list of items. All interactions are treated as positive, and negative instances are randomly selected because of no explicit feedback. 

Additionally, Kariyernet dataset was used for performance evaluation in the experiments, and provided us mutual recommendation. It finds the best candidate list for a job posting, and ranks the most suitable jobs for a candidate. The results show that HR and NDCG evaluation metrices for item recommendation is higher than user recommendation since while the number of job postings is more limited for a candidate, candidate set of a job posting is more wider.

As future work, we will study on processing text-based information of users and items since a great amount of information exists in text. For example, detailed explanation of experiences and qualifications of a candidate, and explanation of a job posting will expand our latent feature space while doing JR task. Also, we are randomly selecting items to be ranked for users, and finding top-n list of these items in this work, but it can recommend irrelevant items to users because of random selection in a real application. Therefore, we aim at realizing candidate generation task before generating top-n list. 

%
%
%
%

\end{document}